\documentclass[submission,copyright,creativecommons,hidelinks, colorlinks=true, allcolors=purple]{eptcs}
\providecommand{\event}{ICLP 2026} 

\let\overleaf=y   

\usepackage{graphicx}
\graphicspath{{../}}

\usepackage{iftex}

\ifpdf
  \usepackage{underscore}         
  \usepackage[T1]{fontenc}        
\else
  \usepackage{breakurl}           
\fi

\title{Modeling Deontic Modal Logic in ASP}
\author{Gopal Gupta$^1$ \qquad Abhiramon Rajasekharan$^1$ \qquad Alexis R. Tudor$^1$ \\ Elmer Salazar$^1$ \qquad Joaquín Arias$^2$
  \institute{$^1$The University of Texas at Dallas\\ Richardson, TX, USA
  \and
  $^2$CETINIA, Universidad Rey Juan Carlos\\Madrid, Spain}
  }
  
\newcommand{\titlerunning}{Modeling Deontic Modal Logic in ASP}
\newcommand{\authorrunning}{Gopal Gupta et al.}

\hypersetup{
  bookmarksnumbered,
  pdftitle    = {\titlerunning},
  pdfauthor   = {\authorrunning},
  pdfsubject  = {EPTCS},               
}

\usepackage{amsmath}
\usepackage{graphicx}
\usepackage{multirow}
\usepackage{xspace}
\usepackage{multicol}
\usepackage{longtable}

\usepackage{xcolor}
\usepackage{adjustbox}
\usepackage{amssymb}
\usepackage{booktabs}

\usepackage{fontawesome5}
\newcommand{\scasp}[1][]{\href{https://ciao-lang.org/playground/scasp.html}{\color{black}\faExternalLink*}}
\ifx y\overleaf
\newcommand{\scaspaux}[3]{%
  \ifthenelse{\equal{#1}{#2}}%
  {\href{#3}{\color{saved}\faExternalLink*}}%
  {}%
}
\renewcommand{\scasp}[1]{%
  \colorlet{saved}{.}
  \scaspaux{#1}{Chisolm1}{https://bit.ly/3GszbCP}%
  \scaspaux{#1}{Chisolm02}{https://tinyurl.com/29zxjfpd}%
  \scaspaux{#1}{asparagus}{https://bit.ly/44rrrcj}%
  \scaspaux{#1}{dog}{https://bit.ly/4kkDY7m}%
  \scaspaux{#1}{kill}{https://bit.ly/3TMX1fx}%
  \scaspaux{#1}{Sartre}{https://bit.ly/3TMFxjo}%
  \scaspaux{#1}{Kant}{https://bit.ly/44Kprx3}%
  \scaspaux{#1}{speed}{https://bit.ly/3IsTr7T}%
  \scaspaux{#1}{car02}{https://tinyurl.com/ycy7xz84}%
  \scaspaux{#1}{updated_ex}{https://tinyurl.com/ycy7xz84}%
  \scaspaux{#1}{car}{https://tinyurl.com/4dyuswu8}%
  \scaspaux{#1}{}{https://ciao-lang.org/playground/scasp.html}%
   \scaspaux{#1}{cabalar1}{https://bit.ly/4nAGF7C}%
   \scaspaux{#1}{cabalar2}{https://tinyurl.com/ybfyv366}%
   \scaspaux{#1}{GGencoding}{https://tinyurl.com/m5by8y3k}%
   \scaspaux{#1}{calm}{https://tinyurl.com/ykx5b6n5}%
   \scaspaux{#1}{car03}{https://tinyurl.com/mkr2cpeu}%
   \scaspaux{#1}{killgo}{https://tinyurl.com/58vpu8hw}%
}
\else
\newcommand{\scaspaux}[3]{%
  \ifthenelse{\equal{#1}{#2}}%
  {\href{#3}{\color{saved}\faExternalLink*}}%
  {}%
}
\renewcommand{\scasp}[1]{%
  \colorlet{saved}{.}
  \scaspaux{#1}{Chisolm1}{https://bit.ly/3GszbCP}%
  \scaspaux{#1}{Chisolm02}{https://tinyurl.com/29zxjfpd}%
  \scaspaux{#1}{asparagus}{https://bit.ly/44rrrcj}%
  \scaspaux{#1}{dog}{https://bit.ly/4kkDY7m}%
  \scaspaux{#1}{kill}{https://bit.ly/3TMX1fx}%
  \scaspaux{#1}{Sartre}{https://bit.ly/3TMFxjo}%
  \scaspaux{#1}{Kant}{https://bit.ly/44Kprx3}%
  \scaspaux{#1}{speed}{https://bit.ly/3IsTr7T}%
  \scaspaux{#1}{car02}{https://tinyurl.com/ycy7xz84}%
  \scaspaux{#1}{updated_ex}{https://tinyurl.com/ycy7xz84}%
  \scaspaux{#1}{car}{https://tinyurl.com/4dyuswu8}%
  \scaspaux{#1}{}{https://ciao-lang.org/playground/scasp.html}%
   \scaspaux{#1}{cabalar1}{https://bit.ly/4nAGF7C}%
   \scaspaux{#1}{cabalar2}{https://tinyurl.com/ybfyv366}%
   \scaspaux{#1}{GGencoding}{https://tinyurl.com/m5by8y3k}%
   \scaspaux{#1}{calm}{https://tinyurl.com/ykx5b6n5}%
   \scaspaux{#1}{car03}{https://tinyurl.com/mkr2cpeu}%
   \scaspaux{#1}{killgo}{https://tinyurl.com/58vpu8hw}%
}
\fi

\usepackage[
textwidth=3.3cm,textsize=footnotesize]{todonotes}
\newcommand{\commin}[1]{\todo[inline]{#1}}

\usepackage{adjustbox}
\usepackage{relsize}
\definecolor{PrologPredicate}{RGB}{0,0,200}
\definecolor{PrologVar}      {RGB}{145,032,039}
\definecolor{PrologComment}  {RGB}{100,170,0}
\definecolor{PrologOther}    {rgb}{0.2,0.2,0.2}
\definecolor{PrologString}   {RGB}{070,120,200}
\usepackage{listings} 
\usepackage{textcomp}
\newcommand{\code}{\lstinline[style=MyInline]}
\lstdefinestyle{MyInline}
{
  basicstyle = \relsize{-0.5}\ttfamily\color{PrologPredicate},
  moredelim = {*[s][]{(}{)}},
  moredelim = {*[s][]{'}{'}},
  moredelim = {*[s][]{\ :-}{.}},
  mathescape=true,
  breaklines = true,
  breakatwhitespace=true,
  upquote = true,
  literate =
  {?-}{{?-\,}}3
  {:-}{{:-\,}}3
  {.=.}{{\,\#=\,}}3
  {.<.}{{\,\#<\,}}3
  {.>.}{{\,\#>\,}}3
  {.=<.}{{\,\#=<\,}}4
  {.>=.}{{\,\#>=\,}}4
}
\lstdefinestyle{MySCASP}
{
    xleftmargin=0.5cm,
    numberstyle=\tiny,
    numbers=left,
    stepnumber=1,
  aboveskip = 0.1em,
  mathescape = true,
  basicstyle = \ttfamily\relsize{-0.5}\color{PrologPredicate},
  basewidth = 0.50em,
  moredelim = {*[s][\color{PrologVar}]{(}{)}},
  moredelim = {*[s][\color{PrologString}]{'}{'}},
  moredelim = {*[s][\color{PrologOther}]{\ :-}{.}},
  commentstyle = \mdseries\color{orange!70!black},
  morecomment=[l]\%,
  literate     =
  {:-}{{\textcolor{orange!70!black}{:-}}}2
  {?-}{{\textcolor{orange!70!black}{?-}}}2
  {.=.}{\#=}2
  {.<.}{\#<}2
  {.>.}{\#>}2
  {.=<.}{\#=<}3
  {.>=.}{\#>=}3
}
\lstdefinestyle{tree}
{
    xleftmargin=0.5cm,
    numberstyle=\tiny,
    numbers=left,
    stepnumber=1,
  mathescape = true,
  basicstyle = \ttfamily\relsize{-.5}\color{PrologString},
  basewidth = 0.45em,
  moredelim = {*[s][\color{PrologString}]{(}{)}},
  moredelim = {*[s][\color{PrologString}]{'}{'}},
  moredelim = {*[s][\color{PrologString}]{\ :-}{.}},
  literate     =
  {.=.}{\#=}2
  {.<.}{\#<}2
  {.>.}{\#>}2
  {.=<.}{\#=<}3
  {.>=.}{\#>=}3
}
\newcommand{\redsection}{\vspace{-1.0em}}
\newcommand{\redsubsection}{\vspace{-0.5em}}

\begin{document}
\maketitle

\begin{abstract}
We consider the problem of implementing deontic modal logic. We show how (deontic) modal operators can be elegantly and directly expressed using default negation (negation-as-failure) and strong negation present in answer set programming (ASP). We propose using global constraints of ASP to represent obligations, prohibitions, and permissions in deontic modal logic. We show that our proposed representation results in the various decades-old paradoxes of deontic modal logic being simply and elegantly resolved. Our method also serves as a means for modeling conditional obligations and conditional prohibitions in knowledge representation.
\end{abstract}


\redsection
\section{Introduction}
Modal logic extends classical logic with modal operators like ``it is necessary that'' and ``it is possible that'' to reason about possibility, necessity, and related modalities. It is also used to formally model statements about what could, must, or ought to be true across possible worlds.
Deontic modal logic is a modal logic that formalizes reasoning about normative concepts such as obligation (\textrm{OB}), permission (\textrm{PE}), and impermissibility (\textrm{IM}). It uses modal operators to express statements like ``It ought to be the case that A'' or ``It is permitted that B''. Deontic logic helps model ethical, legal, and normative systems by capturing how actions are classified as obligatory, permissible, or forbidden.

Modal logics, including deontic logic, have been heavily studied for decades. In this paper, we focus on showing how modal logics can be \textit{directly} represented in answer set programming (ASP), a formalism for knowledge representation and reasoning \cite{Baral,cacm-asp,gelfondkahl}. 
The negation operator used in modal logics is of two types: negation appearing next to a proposition and that appearing next to a modal operator. The former is mapped to  \textit{strong} negation of ASP, and the latter is mapped to \textit{default} negation (also called negation-as-failure \cite{Baral}). We show that this mapping is compatible with the \textit{traditional threefold classification} and the \textit{modal square} (and \textit{deontic square}) \textit{of opposition} in modal logic \cite{seph-deontic-logic}. We discuss how alethic and deontic modal logics can be represented in ASP. In particular, we use ASP's \textit{global constraints} \cite{Baral} for representing obligations, permissibilities, and prohibition. This results in a direct and elegant encoding of deontic logic formulas in ASP. \textit{We show that the paradoxes of deontic logic can be resolved with this representation}. Finally, we give a comprehensive example to illustrate our ideas.

Our main contribution is to show how modal logics, particularly deontic logic, can be directly represented in ASP, and how this representation allows long-standing paradoxes to be elegantly resolved. 
Our work also shows how ASP's \textit{global constraints} can be used for modeling obligations, prohibitions, and permissions in knowledge representation tasks, how conflicting norms can be modeled, and how they can be selectively included or excluded in a programmatic manner while representing knowledge.    

\redsection
\section{Answer Set Programming} 
Answer Set Programming is a popular logic-based paradigm for knowledge representation and reasoning based on the stable model semantics (details about ASP can be found elsewhere \cite{Baral, cacm-asp,gelfondkahl}). ASP supports \textit{negation-as-failure} or \textit{default} negation, denoted \code{not p}, as well as \textit{strong} negation, denoted \code{-p}. Note that \code{not p} is true if we fail to prove \code{p}, while explicit rules containing \code{-p} in their head must be defined to establish that \code{-p} holds, i.e., \code{p} is false. Broadly speaking, ASP consists of three constructs: 
%

\smallskip 
\noindent{\bf 1. Default Rules:} 
Default rules allow us to jump to a conclusion in the presence of incomplete knowledge. Default rules can contain exceptions that prevent us from jumping to the default conclusion. Preferences can also be incorporated in default rules if multiple rules exist. An example is the following: ``normally birds fly, unless they are penguins''.




\begin{lstlisting}
flies(X) :- bird(X), not exception(X).
exception(X) :- penguin(X).
\end{lstlisting}


\noindent{\bf 2. Integrity Constraints (Denials):} 
Integrity constraints in ASP (also called denials) allow us to state the combination of predicates that are collectively inconsistent. As an example, consider a denial that enforces that a dead person cannot breathe:

\begin{lstlisting}
false :- person(X), dead(X), breathe(X).
\end{lstlisting}

\vspace{-0.075in}
\noindent For an object that is not person or not dead, the conjunction is false. If a \code{person} \code{X} is \code{dead},  the predicate \code{breathe/1} should be false for the constraint to hold (the keyword \code{false} is optional). Note that denials are a special case of \textit{odd loops over negation (OLON)} \cite{Baral,cacm-asp}, and the constraint above is really the following:
%
%

\begin{lstlisting}
ic(X) :- person(X), dead(X), breathe(X), not ic(X). 
\end{lstlisting}

\vspace{-0.07in}
\noindent if \code{ic(X)} is true because of other rules, then this OLON rule will be taken out of consideration (due to \code{not ic(X)} in the body being false). Otherwise, the OLON rule amounts to forcing the denial that the conjunction of \code{person(X), dead(X), breathe(X)} should be false. 
It should be noted that in ASP, the conjunction of \code{p} and \code{-p} is always false: for every predicate \code{p}, the denial \code{:- p, -p} is asserted.

\smallskip 

\noindent{\bf 3. Even Loops Over Negation}:
%
Even loops over negation allows us to generate multiple possible worlds. E.g., the rules\ \ \code{g :- not not_g}\ , and\ \ \code{not_g :- not g}\ , lead to two answer sets (possible worlds): one in which \code{g} is true (and \code{not_g} false), and the other in which \code{g} is false (and \code{not_g} true). We can think of \code{g} as an \textit{abducible} goal that is alternately set to true and false in two different possible worlds.

We employ the s(CASP) goal-directed predicate ASP system \cite{scasp} in our work, where the above effect is achieved by simply declaring \code{g} as an abducible via the directive \code{#abducible g}. It generates the even loop over negation described above.
%
Such an abducible can be asserted for any arbitrary predicate. 

\redsection
\section{Alethic Modal Logic in ASP}
\label{alethic}
\redsubsection 

Alethic modal logic is the logic of necessary truths and related notions. Detailed exposition of alethic modal logic is provided elsewhere \cite{seph-deontic-logic}. Alethic modal logic considers six basic modal notions (with $p$ denoting a proposition; see Fig.~\ref{fig:alethic}).
%
%
 It is well known that any of the first four operators above can be used to express the remaining five operators. Fig.~\ref{fig:alethic} shows this denotation, considering that we denote the first one ``it is necessary that $p$'' as $\Box p$, ($\Diamond p$ is generally used to denote ``possible $p$''). 

\begin{figure}[h]
  \centering
  \begin{tabular}{l@{\hskip 1cm}l}
    \toprule
    Modal notion & Denotation\\
    \midrule
    1. it is necessary that $p$     & $\Box p$\\
    2. it is possible that $p$   &$\neg\Box\neg p$ ( $\equiv$  $\Diamond p$ ) \\ 
    3. it is impossible that $p$    & $\Box\neg p$\\                         
    4. it is non-necessary that $p$ & $\neg\Box p$\\                        
    5. it is contingent that $p$    & $\neg\Box p \wedge \neg\Box\neg p$\\   
    6. it is non-contingent that $p$& $\Box p \vee \Box\neg p$              \\
    \bottomrule
\end{tabular}
  \caption{Basic modal notions in Alethic modal logic and their denotation.}
  \label{fig:alethic}
  \vspace{-1em}
\end{figure}


Alethic modal logic satisfies the basic axioms of modal logic shown below:

\begin{tabular}{@{\hskip .5cm}l@{\hskip 10cm}rl}
1. All tautologous formulas\\
2. \textbf{D:} $\Box p \rightarrow \neg\Box\neg p$ ($\Diamond p$)\\
3. \textbf{K:} $\Box (p \rightarrow q) \rightarrow (\Box p \rightarrow \Box q)$ \\
4. \textbf{NEC:} if $\vdash p$ then $\vdash \Box p$
\end{tabular}

\smallskip
\noindent Additionally, Alethic modal logic also satisfies the following two axioms:

\indent\indent $\Box p \rightarrow p$ (if it is necessary that $p$, then $p$ is true). 

\indent\indent $\Box \neg p \rightarrow \neg p$ (if p is impossible, then p is false).

\smallskip
\noindent 
The second premise can be written contrapositively as $p \rightarrow \Diamond p$, i.e., if $p$ holds then $p$ is possible. %
The six modal notions listed above are directly expressible using default negation and strong negation of ASP.
Negation-as-failure or default negation represented as \code{not p} evaluates to true if \code{p} is false or unknown. It evaluates to false if \code{p} is true. Strong negation, denoted \code{-p}, evaluates to false if \code{p} is true or unknown. It evaluates to true, if the falsehood of \code{p} can be explicitly established. Crucially, if \code{p} is totally unknown (i.e., no rules for either \code{p} or \code{-p}), then \code{-p} evaluates to false while \code{not p} evaluates to true.

\begin{figure}
\begin{center}
  \vspace{-0.15in}
  \includegraphics[width=.65\columnwidth]{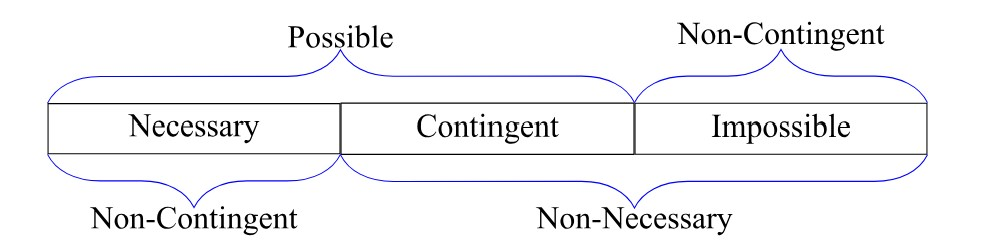}
  \vspace{-0.1in}
  \includegraphics[width=.6\columnwidth]{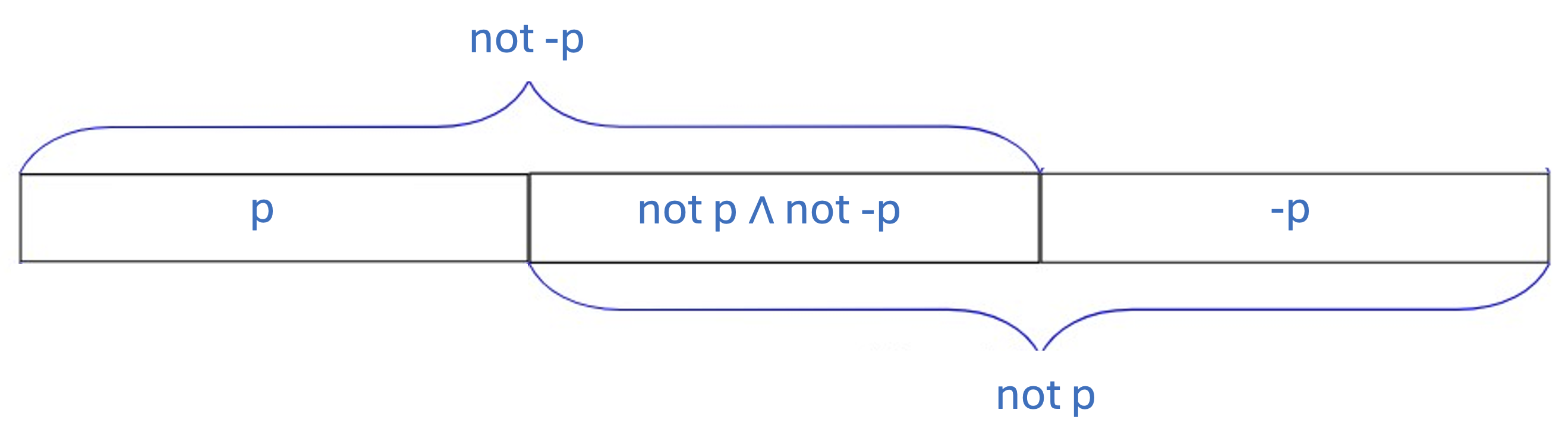}
    \caption{3-fold partition of propositions in Deontic logic compared to ASP notation}
 
    \label{threefold}
    \end{center}
  \vspace{-2em}
\end{figure}

In our ASP representation of modal logic, negation appearing next to a proposition is treated as strong negation and that appearing next to a modal 
operator is treated as default negation or negation-as-failure. Thus, modal logic's $\neg\Box p$ translates to ASP's \code{not p}, while  $\neg p$ translates to \code{-p}. 
Fig.~\ref{threefold} shows the mapping to the 3-fold partition of propositions in alethic modal logic taken from the Stanford Encyclopedia of Philosophy's article on deontic logic by \cite{seph-deontic-logic}. Given a proposition, it is either true (necessary), false (impossible), or unknown (contingent). It is easy to see that default negation (\code{not p}) corresponds to ``impossible or contingent'', while default negation, applied to a strongly negated goal (\code{not -p}), corresponds to ``necessary or contingent''. 




\smallskip 

This observation can be used to program various shades of truth and judgment calls through a combination of negation-as-failure and {strong negation} in ASP \cite{gupta-csr}. For example, given a proposition \code{p}:

\vspace{-.5em}
\begin{enumerate}\itemsep=-2pt
    \item \code{p}: denotes that \code{p} is \textit{unconditionally} true (necessary $p$).
    \item \code{not -p}: denotes that \code{p} \textit{maybe} true (possible $p$).
    \item \code{not p $\wedge $  not -p}: denotes that \code{p} is \textit{unknown}, i.e., no evidence of either \code{p} or \code{-p} (contingent $p$). 
    \item \code{not p}: denotes that \code{p} \textit{may} be false, i.e., no evidence of \code{p} (non-necessary $p$).
    \item \code{-p}: denotes that \code{p} is \textit{unconditionally} false (impossible $p$).
\end{enumerate}

\noindent As noted, the notion ``unconditionally true $p$'' maps to ``necessary $p$'', ``maybe true $p$'' maps to ``possible $p$'', ``unconditionally false $p$'' maps to ``impossible $p$'', ``unknown $p$'' maps to ``contingent $p$'', and ``maybe false $p$'' maps to ``not necessary $p$''. Thus, we can write these modal notions in the following way:

\smallskip

\begin{tabular}{l@{\hskip .1em}r@{\hskip .1em}l@{\hskip 10cm}}
  1. Necessary that $p$: &$\Box p$ &$\ \equiv\ $ \code|p|\\
  2. Possible that $p$: &$\Diamond p$  $\equiv$ $\neg\Box\neg p$ &$\ \equiv\ $ \code|not -p|\\
  3. Impossible that $p$: &$\Box\neg p$ &$\ \equiv\ $ \code|-p|\\
  4. Not necessary that $p$: &$\neg\Box p$ &$\ \equiv\ $ \code|not p|\\
  5. Contingent that $p$: &$\neg\Box p \wedge \neg\Box\neg p$ &$\ \equiv\ $ \code|not p $\wedge $ not -p|\\
  6. Non-contingent that $p$: &$\Box p \vee \Box\neg p$ &$\ \equiv\ $  \code|p $\vee$ -p|\\
\end{tabular}

\medskip

\noindent 
The fact that ASP asserts the denial \code{:- p, -p.} for every proposition \code{p} ensures that axiom \textbf{D} holds for the ASP modeling of alethic modal logic.

{\bf D:} $\Box$\code{p} $\rightarrow$ $\neg\Box\neg$\code{p} = true
~~~~ $\ \ \Rightarrow$ \code{p} $\rightarrow$ \code{not -p} = true
~~~~ $\ \ \Rightarrow\ \ 
$ \code{not p} $\vee$ \code{not -p} = true \hfill

~~~~~~~~~~~~~
$\ \ \Rightarrow\ \ $ 
\code{not (p $\wedge$ -p)} = true 
~~~~~~ $\ \ \Rightarrow$ 
\code{p $\wedge$ -p} = false


\medskip 
\noindent With this observation, the modal square of opposition automatically follows (see Fig. 2 in \cite{seph-deontic-logic}), i,e, the modal square of opposition states that if \code{p} is necessary, it is possible, and if \code{p} is impossible, then it is not necessary. In ASP, if \code{p} is true, then \code{-p} must be false, which implies that \code{not -p} (possible \code{p}) holds; and if \code{p} is impossible, then \code{-p} is true, so \code{p} must be false, and so \code{not p} (\code{p} is not necessary) is true. 


As stated earlier, strong negation captures the notion of impossibility, while negation-as-failure captures the notion of the modal not operator. 
The $\Box$ (necessity) operator disappears due to the  alethic modal logic rule: $\Box p \rightarrow p$.
With the mapping described above, formulas of alethic modal logic can be encoded in ASP and their satisfiability checked. 

\redsection
\section{Deontic Modal Logic in ASP}
\label{opdef}

Next, we show that deontic modal logic can also be modeled in ASP. Deontic logic is the logic of obligation, permission, and prohibition; and like alethic modal logic has six modal notions (Fig.~\ref{fig:deontic}). 
Like in alethic modal logic, Fig.~\ref{fig:deontic} shows that we can express operators 2 through 6 using the OB$p$ operator (denoting $p$ is obligatory).

{\small 
\begin{figure}[h]
  \centering
  \begin{tabular}{l@{\hskip 1cm}l@{\hskip 1cm}l}
    \toprule
    Modal notion & Denotation 
    \\
    \midrule
1. it is obligatory (OB) that $p$    & $\textrm{OB}p$ 
\\                                   2. it is permissible (PE) that $p$   & $\textrm{PE}p \equiv \neg \textrm{OB}\neg p$ 
\\                        
3. it is impermissible (IM) that $p$ & $\textrm{IM}p \equiv \textrm{OB}\neg p$
\\                             
4. it is omissible (OM) that $p$     & $\textrm{OM}p \equiv\neg\textrm{OB}p$ 
\\                               
5. it is optional (OP) that $p$      & $\textrm{OP}p \equiv \neg\textrm{OB} p \wedge \neg\textrm{OB}\neg p$
\\
6. it is non-optional (NO) that $p$  & $\textrm{NO}p\equiv\textrm{OB}p \vee \textrm{OB}\neg p$ 
\\             
    \bottomrule
\end{tabular}
  \caption{Basic modal notions in Deontic modal logic and their denotation.}
  \label{fig:deontic}
  \vspace{-1em}
\end{figure}
}



\noindent Given the similarity of deontic and alethic logic, obligation is referred to as deontic necessity \cite{seph-deontic-logic}. Standard Deontic Logic (SDL) is defined using the following axioms:

\smallskip

\begin{tabular}{l@{\hskip 10cm}}
1. All tautologous formulas\\
2. \textbf{D:} $\textrm{OB}p \rightarrow \neg\textrm{OB}\neg p$ ($\textrm{PE}p$)\\
3. \textbf{K:} $\textrm{OB}(p \rightarrow q) \rightarrow (\textrm{OB} p \rightarrow \textrm{OB} q)$ \\
4. \textbf{NEC:} if $\vdash p$ then $\vdash \textrm{OB} p$
\end{tabular}

\medskip

\noindent However, the deontic analogs of two principles for alethic necessity cited earlier do not necessarily hold:

\indent If OB$p$ then $p$ (if $p$ is obligatory, then $p$
is true).

\indent If IM$p$, then $\neg p$ (if $p$ is impermissible, then $p$ is false).

\medskip 
\noindent This is because obligations can be violated (e.g., we may not pay a parking fine even if we are obligated to do so) and impermissible things do happen (e.g., a car is driven on the wrong side of the road).
We will assume that all deontic formulas used are written as Horn logic programs with deontic operators OB, IM, and PE allowed as qualifiers of subgoals. 
We systematically translate these deontic formulas into ASP rules. We assume that the stable model semantics \cite{Baral} is employed and that the program consisting of these ASP rules can be executed using an answer set programming system, specifically the s(CASP) predicate ASP system (\cite{scasp,gupta-csr}). The semantics of modal logics is given in terms of possible worlds, so modal logic theories can be elegantly represented and reasoned about in ASP. 


On the surface, deontic logic appears similar to alethic modal logic, with obligation being the counterpart of necessity, permissibility of possibility, and impermissibility of impossibility. However, that is not the case, and this superficial similarity has led to many paradoxes in deontic logic \cite{old-deontic-paradox-paper}.

\redsubsection
\subsection{Norms as Global Constraints}

Formalization of deontic logic has been difficult and controversial. Most of the issues arise due to the difficulties in representing norms (obligations, permissibilities, and prohibitions). Quoting from \cite{kow1}:

\begin{quote}
Deontic logic is concerned with representing and reasoning about norms of behavior. However, many authors have denied ``the very possibility of the logic of norms and imperatives''~\cite{hilpinen-mcnamara}. Makinson 
\cite{makinson}, in particular, states that ``there is a singular tension between the philosophy of norms and the formal work of deontic logicians. $\dots$ Declarative statements may bear truth-values, i.e., are capable of being true or false, whilst norms are items of another kind. They assign obligations, permissions, and prohibitions. They may be applied or not, respected or not $\dots$ But it makes no sense to describe norms as true or as false.'' However, \cite{jorgensen}, while acknowledging this distinction between norms and declarative sentences, noted that there are ``inferences in which one or both premises as well as the conclusion are imperative sentences, and yet the conclusion is just as inescapable as the conclusion of any syllogism containing sentences in the indicative mood only''. The resulting conundrum is known as 
\textit{Jørgensen’s dilemma}.
\end{quote} 

\noindent This conundrum can be resolved by considering that norms impose a \textit{constraint} on the accessible worlds. That is, if OB$p$ is true in a world $w$, then \code{p} must hold in all worlds $v$ accessible to $w$. In other words, $p$ \textit{must} be a logical consequence of the theory that represents the world $v$. Viewed as a constraint, the obligation OB$p$ simply states that in any model of the theory representing the world $v$, \code{p} must hold. Similarly, IM$p$ is true only where negation of \code{p} holds, i.e., \code{-p} is a consequence of the theory that represents that world. Finally, PE$p$ is true only where \code{-p} cannot be inferred, i.e., \code{not -p} is a consequence of the theory that represents that world.

As discussed earlier, constraints or denials are an integral part of ASP. An obligations or an impossibility can be expressed through denials in ASP. If OB$p$ (resp. IM$p$) holds in world $w$, then \code{p} (resp. \code{-p}) holds in every world $v$ accessible to $w$. Thus, the obligation \code{p} (OB$p$) can be explicitly enforced in ASP by asserting: \code{:- not p}.
The above denial only admits accessible worlds in which \code{p} holds. Likewise, the impossibility of \code{p} can be explicitly enforced by asserting the denial \code{:- not -p}, and the permissibility of \code{p} can enforced by asserting the denial \code{:- -p.}

Given a rule, \code{p :- q$_1$, q$_2$, $\dots$, q$_n$}, the obligation to enforce it, OB($(q_1, q_2,$ $\dots,~q_n)~\rightarrow~p$), simply states that in every accessible world, we either force the conjunction \code{q$_1$, q$_2$, $\dots$, q$_n$} to be false or we force \code{p} to be true. This translates to enforcing the denial/constraint \code{:- not p, q$_1$, q$_2$, $\dots$, q$_n$.} 
In deontic logic, generally, IM$((q_1, q_2,\dots,~q_n)~\rightarrow~p)$ and PE$((q_1, q_2,$ $\dots,~q_n)~\rightarrow~p)$ are less useful, so we do not consider them here. 

%
%

Given the rule: $q_1, q_2, \dots, q_n \rightarrow \textrm{OB}p$, if \code{q$_1$, q$_2$, $\dots$, q$_n$} are in an answer set, then \code{p} must be forced to be in that answer set as well. Such a rule then translates to the denial \code{:- not p, q$_1$, q$_2$, $\dots$, q$_n$}.
%
%
%
%
%
Likewise, given the rule: $q_1, q_2, $\dots$, q_n \rightarrow \textrm{IM}p$, 
then if \code{q$_1$, q$_2$, $\dots$, q$_n$} are in an answer set, then \code{-p} must be forced to be in that answer set as well. Such a rule, hence, also translates to a denial, in this case: ~~\code{:- not -p, q$_1$, q$_2$, $\dots$, q$_n$}. 
Similarly, for the PE norm: $q_1, q_2, $\dots$, q_n \rightarrow \textrm{PE}p$ translates to the denial \code{:- -p, q$_1$, q$_2$, $\dots$, q$_n$}.

Note that in our approach OB($Q \rightarrow p$) and $(Q \rightarrow OBp)$ both translate to the denial \code{:- Q, not p.} This is because the former states that ``it is required that ($Q$ implies $p$)'', which means that we must either make $Q$ false or $p$ true. This equates to the denial \code{:- Q, not p.} The latter states that we must enforce $p$ if $Q$ holds, which also equates to the denial \code{:- Q, not p.} So coincidentally both map to the same ASP construct, though the difference is that in the former case the value of $Q$ and $p$ considered are the ones in the accessible world, while in the latter case, the value of $Q$ considered is the one in the current world, and the value of $p$ considered is in the accessible world. Thus, the world where $Q$ is evaluated is the differentiator.

Further note that our approach can clearly distinguish between factual detachment and deontic detachment. Deontic detachment stands for $\textrm{OB}p  
\wedge (p\rightarrow \textrm{OB}q)  \rightarrow \textrm{OB}q$, while factual detachment stands for $p 
\wedge (p\rightarrow \textrm{OB}q)  \rightarrow \textrm{OB}q$. The two are distinctly modeled as follows in our approach:

\vspace{-1em}
\begin{multicols}{2}
\begin{lstlisting}
% factual detachment
:- p, not q.
p.
\end{lstlisting}
\begin{lstlisting}
% deontic detachment
:- p, not q.
:- not p.
\end{lstlisting}
\end{multicols}
\vspace{-1.5em}

\noindent In both cases, we accept only those worlds in which the denial \code{:- not q.} (OB$q$) holds. For factual detachment, it is easy to see that \code{q} \textit{must} be true in any acceptable world. Deontic detachment insists that \code{p} must be true (through other means) in any acceptable world, which implies that \code{q} \textit{must} hold as well. 

\redsubsection
\subsection{ASP Representation 
}

To formalize the above ideas, we consider propositional Horn formulas, extended with modal operators.  The syntax of these formulas is shown below. 

[\textbf{M}](([\textbf{M}]$q_1$ $\wedge$ [\textbf{M}]$q_2$$\wedge$ ...$\wedge$ [\textbf{M}]$q_n$) $\rightarrow$ $p$)  ~~~~~~~~...~~(i)

([\textbf{M}]$q_1$$\wedge$ [\textbf{M}]$q_2$$\wedge$ ...$\wedge$ [\textbf{M}]$q_n$) $\rightarrow$ [\textbf{M}]$p$ ~~~~~~~~~~~~...~~(ii)

\smallskip
\noindent where \textbf{M} stands for one of the modal operators: OB, IM, or PE. The modal operator is optional, indicated by square brackets. The antecedent in each rule may be empty (i.e., the rule body maybe just \textit{true}). We do not consider nested modal operators as is customary in modal logic literature (see the beginning of Section 2 in the Stanford Encyclopedia of Philosophy article \cite{sep-logic-modal} on modal logic for discussion on this). Also, we don't permit the modal operators to be negated, since a negated modal operator can be expressed directly using another modal operator (e.g., $\neg$PE$p$ = IM$p$). We translate these modal logic programs into answer set programs via transformations discussed above and formalized below:

\vspace{-0.05in}
\noindent
\begin{enumerate}
\item First consider that the antecedent is empty (\textit{true}). If the rule has a modal operator in front of the rule: OB$p$ translates to \code{:- not p}; IM$p$ translates to \code{:- not -p}; and PE$p$ translates to \code{:- -p}.
If there is no modal operator, then the rule $true \rightarrow p$ simply translates to an ASP fact \code{p}.

Note that if we assume \code{p} and \code{-p} to be abducibles, then the denial corresponding to OB$p$ will admit only one world, \code|{p}|, the denial corresponding to IM$p$ will also admit only one world, \code|{-p}|, while the denial corresponding to PE$p$ will admit two possible worlds, \code|{p}| and \code|{}| which is consistent with \code{p} being possible. 

\item Now, let's examine the case where the antecedent is non-empty.
If a subgoal in the antecedent is not qualified by a modal operator, it stays as is.
If it is qualified, then the subgoal OB$q_i$ in the antecedent simply translates to \code{q$_i$}, since \code{q$_i$} will trigger the rule only if \code{q$_i$} is present in the answer set through other means. However, we should have a rule with OB$q_i$ in its head, since the condition under which the obligation for $q_i$ must hold must be given. Similarly, IM$q_i$ in the antecedent translates to \code{-q$_i$}, and there must be a rule present with IM$q_i$ in its head. Likewise, PE$q_i$ in the antecedent translates to \code{not -q$_i$} and there must be rule with PE$q_i$ present in its head. 

We essentially exploit the correspondence between modal logic and ASP discussed earlier regarding how the various shades of truth discussed in Section \ref{alethic} can be represented with default and strong negation. Also, in the last two cases (IM$q_i$ $\equiv$ \code{-q$_i$} and PE$q_i$ $\equiv$ \code{not -q$_i$}), we have to ensure that we have appropriate rules to define \code{-q$_i$}.

\item
After the above steps, all modal operators disappear from the \textit{antecedents} of the formulas, and we obtain formulas of the form $Q \rightarrow [\textbf{M}]p$ or [\textbf{M}]($Q \rightarrow p$), where $Q$ is a conjunction of literals with no modal operators, and these literals are now possibly qualified by default negation (\code{not q$_i$}), or strong negation (\code{-q$_i$}), or their combination (\code{not -q$_i$}). These formulas are next translated as follows:

\vspace{-.5em}

\begin{enumerate}
\item  $Q \rightarrow \textrm{OB}p$ is translated to the  denial 
\code{:-not p, Q}.
\item $Q \rightarrow \textrm{IM}p$ is translated to the denial 
\code{:-not -p, Q}.
\item  $Q \rightarrow \textrm{PE}p$ is translated to the denial 
\code{:- -p, Q}.
\item OB($Q \rightarrow p$) is translated to \code{:-not p, Q.} (Note: IM($Q \rightarrow p$) and PE($Q \rightarrow p$) are disallowed).
\end{enumerate}

\item Finally, a Horn formula with zero positive literals,
[\textbf{M}]$q_1$ $\wedge$ [\textbf{M}]$q_2$$\wedge$ ...$\wedge$ [\textbf{M}]$q_n$ $\rightarrow false$,
is also processed in the manner described above, giving us a denial. If we wish to distinguish this case from the others, we may equivalently view the resulting clause as an ASP query. However, only one query formula is permitted at a time. 

\end{enumerate}

\noindent
\textbf{Example:} Consider the following deontic logic formulas (on the left) and the corresponding ASP encoding (on the right):

\vspace{-1em}
\begin{multicols}{2}
  1. \textrm{OB}$go$

  2. OB($tell \leftarrow go$)

  3. $\textrm{IM} tell \leftarrow \texttt{-}go$

  4. go
\begin{lstlisting}
:- not go.
:- go, not tell.
:- -go, not -tell.
go.
\end{lstlisting}
\end{multicols}

\vspace{-1em}
\noindent
If we assume \code{tell} and \code{-tell} to be abducibles, we will obtain a single possible world which contains \code{go}, i.e., the model \code{$\{$go,  tell$\}$}.

\medskip 

\noindent\textbf{Example}: If you are required to wear a hazmat suit (OB$hazmat$) [in hazardous area]
\textit{and} the alarm goes off (\textit{alarm}), then it is obligatory to evacuate (OB\textit{evacuate}). The modal formula and ASP code is below: 

\begin{minipage}{.49\linewidth}
  OB$evacuate$ $\leftarrow$ OB$hazmat \land alarm$. 

\vspace{-.2em}
\textrm{OB}$hazmat$ $\leftarrow$ $hazardous\_area$.
\end{minipage}
\begin{minipage}{.49\linewidth}

\begin{lstlisting}
:- hazmat, alarm, not evacuate.
:- hazardous_area, not hazmat.
\end{lstlisting}
\end{minipage}
\vspace{-0.5em}

\noindent In this scenario, if you are in a hazardous area, then a hazmat suit is mandatory, and then if the alarm goes off, you must evacuate. However, if you are in a non hazardous area, and the alarm goes off, then you don't have to evacuate. The ASP code correctly realizes this logic.


\medskip 
\noindent 
\textbf{Theorem}: SDL axioms (\textbf{D}, \textbf{K}, \textbf{NEC}) are satisfied in the ASP representation of norms as denials.

\vspace{.5em}
\noindent 
\textbf{Proof:}
\vspace{-.5em}
\begin{description}
  \item[\textbf{D:}] Axiom D states that $\textrm{OB}p \rightarrow \neg\textrm{OB}\neg p$ is true. That is, $\neg\textrm{OB}p \vee \neg\textrm{OB}\neg p$ holds. Thus, $\neg(\textrm{OB}p \wedge\textrm{OB}\neg p)$ holds. In our ASP encoding, \code{p $\wedge$ -p} being false is ensured because in ASP, for every proposition or predicate \code{-p}, the denial \code{:- p, -p} is asserted. 

  \item[\textbf{K:}] Given $\textrm{OB}(p \rightarrow q) \rightarrow (\textrm{OB} p \rightarrow \textrm{OB} q)$, its antecedent, $\textrm{OB}(p \rightarrow q)$, is translated as \mbox{\code{:- not q, p}}.  
    $\textrm{OB} p$, is translated as \code{:- not p}. It is easy to see that any possible world in which the two denials hold must necessarily have \code{q} hold, i.e., \textrm{OB}$q$ represented as the denial \code{:- not q} holds.
  \item[\textbf{NEC:}]  The necessitation axiom holds, since if any tautology $\tau$ holds in all worlds $w$, i.e., $\vdash \tau$, then \code{:- not $\tau$} holds trivially in all worlds $w$   also, i.e., $\vdash \Box \tau$ holds. \hfill $\Box$
  \end{description} 

\noindent Note that NEC has historically posed a problem as it requires that every tautology be obligatorily enforced. Since OB$p$ becomes a denial (\code{:- not p.}) in our approach, we do not require that the negation of every tautology be explicitly asserted as a denial. 

It is important to note that denials in ASP only perform checks; they cannot draw new conclusions as rules with a non-empty head can do. Any possible world that does not have \code{p} inferred from other rules will not be allowed to exist if the denial \code{:- not p} is applied. This, essentially, precludes the axioms ``if OB$p$ then $p$'' and ``if IM$p$ then $\neg p$'' from holding. Recall that the analogs of these two axioms hold for alethic modal logic. 
This realization is crucial for resolving the ``conundrum'' mentioned above and for resolving the various paradoxes that deontic logic suffers from \cite{seph-deontic-logic}.


%

\redsubsection
\subsection{Preempting Obligations and Impermissibilities}


As noted earlier, obligations may be violated and impermissible things do happen. In such a case, if a violation occurs, then the denial that represents that obligation must be preempted (dropped). Likewise, if a prohibition is breached, then the denial representing the prohibition must be dropped. This can be modeled in ASP quite elegantly because a global constraint can also be represented as an odd loop over negation rule \cite{Baral}, i.e., a rule of the form \code{q :- B, not q} where \code{B} represents a conjunction of goals. 
Under the stable model semantics, if there is a proof of \code{q} through other rules, then the above OLON rule is taken out of consideration and does not contribute to the answer set. However, if there is no alternative proof for \code{q}, then \code{B} \textit{must} be false in every possible world (answer set).
The idea of representing obligations, impermissibilities, and permissibility as OLON rules can be used to elegantly resolve the paradoxes of deontic logic.

Given an obligation OB$p$, and a possible condition \code{c} that results in its violation, then instead of encoding the obligation as \code{:- not p}, we encode it as the OLON rule \code{c :- not p, not c}.
Where, if \code{c} is false, then the denial forces \code{not p} to be false, i.e., \code{p} to be true. While, if \code{c} holds (inferred through some other rules), then the above denial is ignored due to \code{not c} in its body. Thus, the occurrence of a condition that violates an obligation results in that obligation being dropped (similar to what a human would do in such a situation).
Similarly, for impermissibilities, if an impermissible situation \code{q} (IM$q$) can be over-ridden because of a condition \code{c}, we encode it as the OLON rule \code{c :- not -q, not c}. 

This could be even further simplified by representing the obligation as \code{:- not p, not c.}, where \code{c} is the condition under which the obligation is to be dropped. It is easy to see that if \code{c} is true due to other rules, the denial is not enforced.

\redsection
\section{Resolving the Paradoxes of Deontic Logic}
\label{contrary}

We show how various paradoxes of SDL  can be  resolved in our ASP representation, described above. Chief among these is the contrary-to-duty paradox that we consider in detail next. 
Linked s(CASP) programs, provided next to each example, can be executed by invoking the query \code{?- true}.

\redsubsection
\subsection{Contrary-to-duty Paradox}
\label{ctdp}

Also known as the Chisholm's Paradox, introduced by Chisholm in 1963 \cite{ronnedal2}, an example of the contrary-to-duty paradox is the following (on the left is the narrative and on the right its representation in SDL):

\vspace{-1em}
\begin{longtable}{l@{\hskip 1.5cm}l}
  1. It ought to be that Jones goes to assist his neighbors.
  &
    \textrm{OB}$go$\\[-2pt]
  2. It ought to be that if he goes, he tells them he is coming.
  &
    OB($go \rightarrow tell$)  \\[-2pt]
  3. If Jones doesn't go, he ought not tell them he is coming.
  &
    $\neg go \rightarrow \textrm{OB}\neg tell$\\[-2pt]
  4. Jones doesn't go.
  &
    $\neg go$\\  
\end{longtable}

\vspace{-1.25em}
\noindent One can see---and as discussed extensively in the literature---that both OB$tell$ and OB$\neg tell$ can be inferred leading to a contradiction, a paradox. 
Under our proposal, the above is represented in ASP as follows: \hfill \scasp{Chisolm1}


\begin{lstlisting}
-go :- not go, not -go.
:- go, not tell.
:- -go, not -tell.
-go.
\end{lstlisting}


\vspace{-.25em}
\noindent The OLON rule, in line 1, says that Jones must go, unless \code{-go} is proved true through other rules. Line 2 says that in any acceptable world, \code{go} and \code{tell} must be true together. Similarly, line 3 says that \code{-go} and \code{-tell} must be true together. Note that lines 1-3 above simply state the obligations as constraints. Whether Jones goes or not, and whether he tells or not, is completely independent of these constraints. 
So we can now imagine worlds where formulas in lines 1, 2, and 3 are true, and we can choose whatever truth values we wish for \code{go} and \code{tell} by declaring appropriate facts. Only two scenarios will result in an acceptable world: first when both \code{go} and \code{tell} are true, and second when both \code{-go} and \code{-tell} are true. 

Note that in the s(CASP) system, all the possibilities can be generated by declaring \code{go}, \code{-go}, \code{tell}, and \code{-tell} as abducibles. The resulting program can be run on s(CASP) and the various possible worlds obtained through appropriate queries. 
%
%
Given the program in lines 1-4 above, the only acceptable world will be where \code{-tell} is present (i.e., Jones does not tell). By virtue of line 4, \code{-go} will of course be present in this world (Jones does not go). Thus, the constraint representation of norms, where a constraint may be preempted, avoids Chisholm's paradox. 
Given lines 1-3 and the abducibles for \code{go, tell, -go, -tell}, we will find that their are only two admissible worlds \code|{-tell,  -go}| and \code|{tell,  go}|. In the former case, \code{-go} being true will preempt the constraint on line 1.  
%
For a clearer understanding, we can make preemption of the obligation explicit: \hfill \scasp{killgo}

\begin{lstlisting}
ignore_obligation :- not go, not ignore_obligation.
:- go, not tell.
:- -go, not -tell.
ignore_obligation :- -go.
-go.
\end{lstlisting}

\vspace{-.25em}
\noindent It may appear that we can use a simpler alternative formulation where the OLON constraint in line 1 is rewritten as the following, which states that Jones must go, if there is no evidence that he did not go:

~~~\code{go :- not -go}.

\noindent 
However, this encoding suggests that if we don't have any information regarding whether Jones went or not (i.e., neither \code{go} or \code{-go} is known), 
then our theory would infer that he did go, even though we are uncertain if he went or not.
Essentially, an obligation or impermissibility should translate into a denial (global constraint), they should not be used for drawing conclusions. \hfill \scasp{Chisolm02}

\begin{lstlisting}
go :- not -go.
:- go, not tell.
:- -go, not -tell.
tell.
\end{lstlisting}

\vspace{-.5em}
\noindent Other similar paradox relating to strengthening the antecedent \cite{christian-strasser} are similarly resolved. Consider:

\vspace{-.75em}
\begin{enumerate}\itemsep=-2pt
\item Don’t eat with your fingers. ~~~~ \textrm{IM}\textit{fingers}

\item If served asparagus, you must eat it with your fingers. ~~~~ \textit{asparagus} $\rightarrow$ \textrm{OB}\textit{fingers}

\item You are served asparagus.~~~~ \textit{asparagus}
\end{enumerate}

\vspace{-.25em}
\noindent In our formalism, the three formulas will be encoded as: \hfill \scasp{asparagus}

\begin{lstlisting}
asparagus :- not -fingers, not asparagus.  
:- asparagus, not fingers. 
asparagus.
\end{lstlisting}

\vspace{-.25em}
\noindent With this encoding, if \code{asparagus} is served, \code{fingers} must be true. If \code{asparagus} is not served (e.g., the third line is replaced with \code{apple}), \code{-fingers} will be enforced. 

Many complex solutions have been suggested in the literature in the last 50 years to resolve the contrary-to-duty paradox \cite{christian-strasser,ronnedal,prakken-sergot-dyadic} (see \cite{ronnedal2} for a survey of the approaches). These include making the `ought' context sensitive by differentiating between \textit{factual} and \textit{deontic} detachments \cite{seph-detachment}, using dynamic semantics to account for changes in obligations over time \cite{castaneda,vaneck} using dyadic deontic logic \cite{hansson,prakken-sergot-dyadic}, and employing the concepts of sanctions, ordering of models, and abductive logic programming \cite{kow1}, etc. Since humans can deal with deontic paradoxes easily, their resolution in a logical framework should also be simple and elegant. The solutions suggested in the literature are quite complex (see for example proposal in \cite{christian-strasser}). We believe that complexities are significantly reduced by treating an obligation as a global constraint (denial) on the accessible worlds, as illustrated above. 

One of the approaches to resolving the paradox is to use dynamic semantics to account for change in obligations over time  \cite{castaneda}. There have been many attempts to try to solve Chisholm's paradox by carefully distinguishing the time points of the obligations \cite{thomason81b}. As discussed elsewhere \cite{prakken-sergot}, there are versions of the paradox that cannot be solved by analyzing obligations over time, such as the following: (i) It ought to be the case that there are no dogs. (ii) It ought to be the case that if there are no dogs, then there are no warning signs.
(iii) If there are dogs, then it ought to be the case that there are warning signs.
(iv) There are dogs.
%
%
\noindent In our framework, we simply model this  as: 
\hfill \scasp{dog}

1. \code{dog :- not -dog, not dog.}~~~~~~~~~~~~~ 2. \code{:- -dog, not -warning_sign.}

3. \code{:- dog, not warning_sign.}~~~~~~~~~~~~~~4. \code{dog.}

\noindent The query \code{?-warning_sign} posed in s(CASP) will yield the answer set \code|{dog,  warning_sign}|. If we replace line 4 with \code{-dog} and query \code{?- -warning_sign}, the answer obtained is 
\code|{-dog, -warning_sign}|. (Note that both \code{warning_sign} and \code{-warning_sign} will have to be declared as abducibles.) 

\redsection
\subsection{Resolving Other Paradoxes of Deontic Logic}


The contrary-to-duty or Chisholm's paradox has received the most attention. According to McNamara and Van De Putte \cite{seph-deontic-logic}, ``Chisholm’s paradox, and similar cases, were the booster rocket that provided the escape velocity deontic logic needed ...''
There are many other paradoxes in deontic logic that arise due to Jørgensen’s dilemma. These include: Forrester's paradox, Sartre's Dilemma, Ross's paradox, the Good Samaritan Paradox, Kant's Law puzzle, among others. These paradoxes can be  resolved in our representation as well. The description of their resolution is given in Appendix \ref{sec:appA}
due to lack of space. 



\redsubsection
\section{A Comprehensive Example}
\label{sec:carexample}
\redsubsection 

Consider a more complex example taken from \cite{seph-deontic-logic}, which we have slightly modified for clarity of illustration.
{\it John is obligated to return his friend Smith's car, and the least he can do is return it by noon with the same level of battery charge it had when he borrowed it. It is permissible to not maintain the same battery level if the car needs a battery change.}
This example also illustrates how secondary obligations (``least one can do'') can also be elegantly modeled in our approach. 
If the primary obligation is met, there may be secondary obligations to fulfill. 
The main obligation in this example is to return the car (whether or not by noon or with a properly charged battery). If the primary obligation is met, we demand that the secondary obligations (return before noon, return with appropriate battery level) are also fulfilled. 

\begin{lstlisting}
OB car_returned :- borrowed_car. 
OB returned_car_before_noon :- borrowed_car, car_returned.
OB battery_level_ok :- borrowed_car, car_returned,  IM ignore_battery_level.
IM ignore_battery_level :- -needBatteryChange.
\end{lstlisting}


\vspace{-0.05in}
\noindent Note that in rule 3, IM \code{ignore_battery_level} is a replacement for $not$ PE \code{ignore_battery_level}. It captures the exception to enforcing battery charge level if the battery needs changing. We must add the knowledge then that we \textit{cannot} ignore battery's charge level if the battery does not need changing.  
%
%
%
%
%
%
The representation in ASP, shown in Fig.~\ref{fig:carpropositional}, is reasonably straightforward, given the earlier discussion. 
This code can be executed at \url{https://tinyurl.com/33f4dreb}.
Notice that the three obligations represented by the three OLON rules can be preempted by events such as a car being wrecked (so cannot be returned), the borrower becoming sick (so returns car after noon deadline), the borrower being financially broke (no money to charge battery), etc. 
Additional knowledge is needed to state that if a car is returned before noon, then the car was returned, and car returned means the car was borrowed. 

Our work also provides insight into knowledge representation methods. We can model restrictions and invariants as constraints/denials in ASP, e.g., \code{:- c.} 
These constraints/denials can be given ``names'' by turning them into explicit OLON rules, e.g., \code{cname :- c, not cname}.  Separate logic can then be developed to determine when a constraint is to be applied, in an \textit{elaboration-tolerant} manner. That is, we can enforce the constraints \textit{conditionally}, allowing us to realize conditional norms. The extended example above illustrates this well.

\vspace{-0.1in}
\begin{figure}[h]
  \vspace{.25em}
  \begin{multicols}{2}
\begin{lstlisting}[basewidth=.44em]
% If the car can be returned before noon,
% it can be returned.
car_returned :- returned_car_before_noon.
% car can be returned, only if borrowed.
:- car_returned, not borrowed_car.
% Returning the car obligation as an OLON rule.
fail_to_return_car :- 
    not car_returned, borrowed_car,
    not fail_to_return_car.
% Secondary obligation to return before noon.
fail_to_return_by_noon :- 
    not returned_car_before_noon,
    car_returned, borrowed_car,
    not fail_to_return_by_noon.
% ... to return with a battery good.
fail_to_return_ok_battery :- 
    not battery_level_ok, car_returned,
    borrowed_car, -ignore_battery_level,
    not fail_to_return_ok_battery.
battery_level_ok :- same_battery_level.
%Handling permission: can't ignore battery
%level if battery does not need changing
:- not -ignore_battery_level, -needBatteryChange.
-ignore_battery_level :- -needBatteryChange..
% If the car is wrecked, we can't return it.
fail_to_return_car  :- borrowed_car, wrecked_car.
% If sick, you can't return the car by noon.
fail_to_return_by_noon :- borrowed_car, sick.
% If you fail to return the car, you fail
% to return it by noon.
fail_to_return_by_noon :- fail_to_return_car. 
% If broke, you can't charge the battery.
fail_to_return_ok_battery :- 
    borrowed_car, financially_broke, 
    not needBatteryChange.
% If you fail to return the car, then you
%  failed to return with an ok battery.
fail_to_return_ok_battery :- fail_to_return_car.
\end{lstlisting}
  \end{multicols}
  \vspace{-0.15in}
  \caption{ASP Encoding of Car-borrowing Example}
 \label{fig:carpropositional}
  \vspace{-0.1in}
\end{figure}

\redsection
\section{Related Work}

\redsubsection
There is significant work on resolving paradoxes of deontic logic going back decades (see \cite{old-deontic-paradox-paper}). To our best knowledge, none 
model deontic operators as constraints that the accessible worlds must satisfy. 
Many researchers have modeled deontic logic with ASP; however, unlike us, none map it \textit{directly} to ASP. In contrast, we showed how the two types of negation of modal logic map to default and strong negation of ASP. In essence, we showed that ASP is ideal for modeling and implementing alethic and deontic modal logics. Modeling other modal logics is part of future work. 
%
Several works use non-monotonic reasoning to attempt to resolve the paradoxes  \cite{horty,kow2}. 
These efforts rely \textit{indirectly} on the connection between modal operators and default and strong negations.  Abductive logic programming (ALP) has also been proposed for this purpose \cite{kow1,kow2}, which is closely related to ASP. 
The ALP-based approach orders the possible worlds according to a preference relation, then select the best one to avoid the paradox. 
In our approach, this preference is encoded in the answer set program itself. 

Dyadic deontic logic, which introduces a \textit{conditional operator}, has been proposed \cite{prakken-sergot-dyadic} to handle the paradox.
Here, instead of just saying ``You ought to do A,'' we say ``If B is the case, you ought to do A'', resulting in a more nuanced representation of obligations, including those that arise as a consequence of violating other obligations. 
Our method is simpler and can be thought of as taking the context into account to disable constraints that represent obligations in case they are violated. Anderson \cite{anderson} has introduced the notion of sanctions, a specialized propositional constant (often denoted \textit{s} for sanction or \textit{V} for violation) that represents a ``bad thing'' or ``normative demand violation'' which allows deontic logic to be reduced to alethic modal logic. Anderson proposed that standard deontic operators can be defined by treating an obligation as a condition that, if failed, \textit{necessarily} leads to the sanction. In contrast, in our approach, we don't need to introduce the notion of a sanction or violation. The problems with Anderson's approach (and a similar approach by Kanger \cite{kanger}) are well documented \cite{seph-deontic-logic}, e.g., certain obligations such as ``be kind to strangers'' do not have a sanction attached. Our approach does not suffer from these limitations.

In other efforts, 
\cite{govern} has developed a domain-specific language L4 for coding deontic logic. L4 is mapped to defeasible rules, and an ASP meta-interpreter is  used to implement it. 
Alviano et al \cite{alviano} show how arbitrary modal logic formulas can be evaluated using ASP.
Our approach, in contrast, maps deontic logic directly to ASP. Cabalar et al.
\cite{cabalar-deontic} and Hatschka et al. \cite{eiter-deontic} also attempt to realize deontic logic with ASP. Hatschka et al. \cite{eiter-deontic} employ soft constraints to prioritize models (similar to \cite{kow1}). In their approach, various deontic modal operators are \textit{(meta-)interpreted} using ASP, while we \textit{map} them \textit{directly} to ASP. 
%
Cabalar et al. 
introduce predicates corresponding to modal operators. They use a weakened Axiom \textbf{D} called \textbf{wD}: $\neg$(O$p$ $\wedge$ F$p$ $\wedge$ not$~p$ $\wedge$ not $\neg p$). 

Cabalar et al. translate deontic logic formulas to ASP using the operators: O$p$, -O$p$, F$p$, -F$p$, not O$p$, not -O$p$, not F$p$, and not -F$p$, where O$p$ stands for $p$ being obligatory and F$p$ for $p$ being forbidden. 
In Cabalar et al's formulation, F$p$ and O$p$ are allowed to be in an answer set together. However, in such a case, their axiom wD states that $p$ cannot be unknown, i.e., not$~p$ $\wedge$ not$-p$ must be false, i.e., if there is a conflict, then we must choose a truth value for $p$.
%
In contrast, we map the F and O operators to ASP code and do not resort to weakening of Axiom \textbf{D}. In particular, rather than modeling norms using explicit operators as such O$p$, F$p$, etc., we model them as integrity constraint.

\vspace{-0.05in}

\redsection
\section{Conclusion}
\redsubsection

In this paper, we showed how modal logics can be directly realized in ASP. We showed how negation in modal logic maps to the two forms of negation---default and strong---present in ASP. We showed how the obligation, prohibition, and permissibility operators of deontic modal logic elegantly map to conditionally enforced global constraints (denials) of ASP. We also showed how the contrary-to-duty paradox of deontic logic can be  resolved using our approach. We also presented an extended example. Conditional enforcement of constraints is also quite useful in knowledge representation, as our extended example about the obligation to return a borrowed car shows. 

\noindent\textbf{Acknowledgments:}
Authors are grateful to the ICLP anonymous referees for helpful feedback. The work is supported by
project COSASS (PID2021-123673OB-C32) funded by MCIN/AEI/10.13039/501100011033; 
project EVASAI (PID2024-158227NB-C32) funded by MICIU/AEI/10.13039/501100011033 and FEDER, EU;
and by a research gift from Nexco Corp.

\vspace{-1.5em}

\bibliographystyle{eptcs}
\bibliography{aaai2026,scasp}  



\newpage
\appendix

\redsection
\section{Resolving Other Paradoxes of Deontic Logic}
\label{sec:appA}

While the contrary-to-duty or Chisholm's paradox has consumed much attention, there are many more paradoxes in deontic logic. We show how some of these paradoxes are resolved in our formulation based on conditionally enforced global constraints.

\redsubsection
\subsection{Forrester's Paradox}   

Forrester's paradox is a variation of the contrary-to-duty paradox and is stated as follows: 

1. It is forbidden for a person to kill, i.e., \textrm{OB}$\neg kill$;

2. But if a person kills, he ought to kill gently, i.e., $kill \rightarrow \textrm{OB}kill\_gently$;

3. If a person kills gently, then the person kills, i.e., $kill\_gently \rightarrow kill$

\noindent Suppose that Smith kills Jones. Then he ought to kill him gently. But, by the OB-RM theorem of standard deontic logic $((p\rightarrow q) \rightarrow (\textrm{OB} p \rightarrow \textrm{OB} q))$, Smith ought to kill Jones, which contradicts the first obligation, that Smith ought not to kill Jones. These 3 statements can be encoded as: 
\hfill \scasp{kill}

\begin{lstlisting}
kill :- not -kill, not kill.
:- kill, not kill_gently.
kill :- kill_gently.
\end{lstlisting}

\noindent With this encoding, in any possible world where \code{kill} is true, \code{kill_gently} will also be true. So it is not possible for Smith to kill, without killing gently. Additionally, the above theory admits possible worlds where \code{-kill} holds. 

\redsubsection
\subsection{Sartre's Dilemma}

We similarly resolve Sartre's dilemma. 

1. Join the French resistance. i.e. $\textrm{OB}join$; 

2. Stay at home and look after his aged mother. i.e. $\textrm{OB}stay$; 

3. Joining and staying are incompatible. i.e. $\neg(join \wedge stay)$.

The ASP encoding is as follows. It states that staying preempts joining and vice versa. \hfill \scasp{Sartre}

\begin{lstlisting}
join :- not stay, not join.
stay :- not join, not stay.
:- stay, join.
\end{lstlisting}

If we assign all possible combinations of truth values to \code{join} and \code{stay}, then there are only two worlds possible: \code|{join, not stay}| and \code|{stay, not join}|.
This paradox is similar to \textbf{Kant's Law Puzzle} \cite{seph-kant-puzzle}:

1. I'm obligated to pay you back \$10 tonight.
\vspace{-.25em}

2. I can't pay you back \$10 tonight (e.g., just gambled away everything). 

\vspace{.25em}
\noindent encoded as: \hfill \scasp{Kant}

\begin{lstlisting}
broke :- not pay, not broke.
:- broke, pay. 
broke.
\end{lstlisting}

\noindent There is implicit knowledge that if a person is broke, they cannot pay, which is encoded as a constraint. That is, in any possible world entailed by these formulas, \code{broke} and \code{pay} cannot be true together (line 2 above). However, if \code{broke} is false, \code{pay} must be true due to line 1 above. 

\redsubsection
\subsection{Ross's Paradox}

We next consider Ross's Paradox. Ross's paradox arises because of or-introduction. An example is the following (we have modified it slightly for clarity in explanation):  

1. It is obligatory that the letter is mailed. 

2. If the letter is mailed, then the letter is posted or the letter is burned. 

3. Therefore, it is obligatory that the letter is posted or the letter is burned. 
\smallskip

\noindent This is encoded as: 

1. $\textrm{OB}mail$

2. $mail \rightarrow post \vee burn$. 

\noindent Therefore $\textrm{OB}(post \vee burn)$. The paradox arises due to the or-introduction rule: if the statement ``to mail a letter, we post it'' is true, then ``to mail a letter, we post it or burn it'' is also true. However, it should be noted that ``to mail a letter, we post it or burn it'' is true by virtue of ``to mail a letter, we post it''. Essentially, we cannot have a model where ``to mail a letter, burn it'' holds by itself (unless we allow burning the letter as a method of mailing). 
In our formalism, this will be encoded as: 

\begin{lstlisting}
:- not mail.
post :- mail.
burn :- mail.
\end{lstlisting}

\noindent The third rule seems to have been just foisted upon with no basis. In the real world, where we are representing factual knowledge, we cannot simply introduce a disjunction without a basis. This is especially true when we represent knowledge using ASP. Therefore Ross's paradox is really not a paradox.

\redsubsection
\subsection{The Good Samaritan Paradox}

The Good Samaritan paradox states that: (i) It ought to be the case that Jones helps Smith who has been robbed; (ii) If Smith has been robbed and Jones helps Smith, then Smith has been robbed.
(iii) Therefore, it ought to be the case that Smith has been robbed. i.e. \textrm{OB}$(rob \wedge help)$, $(rob \wedge help) \rightarrow rob$. Therefore, \textrm{OB}$rob$. Translating this into ASP, per our encoding, we get the following (Note that \textrm{OB}$(rob \wedge help)$ translates into (1) and (2) below, while $(rob \wedge help) \rightarrow rob$ translates to (3)):

\begin{lstlisting}
:- not rob.        % ...(1)
:- not help.       % ...(2)
rob :- rob, help.  % ...(3)
\end{lstlisting}

This encoding shows that the Good Samaritan paradox is no longer a paradox. In any world in which Smith is robbed, Jones helps him. Given the constraint (1) and (2), a world in which only \code{rob} is true and \code{help} is not cannot exist, given that an obligation is interpreted as a constraint on the world. 


Note that the statement ``It ought to be the case that Jones helps Smith who has been robbed'' can be interpreted as ``If Smith is robbed, then Jones ought to help him'', which will be encoded as 

\begin{lstlisting}
:- rob, not help.
rob :- rob, help.
\end{lstlisting}

\vspace{-0.1in}
\noindent The argument made earlier still applies, and again, arguably, the Good Samaritan paradox is not really a paradox. 

\newpage

\redsection
\section{Narratives for the Car Borrowing Example}

The approach we presented in the paper for propositional deontic modal logic also applies to predicate deontic logic. We present a more general encoding of the car-borrowing example in s(CASP) in which predicate answer set programs can be encoded.  

\begin{figure}[!ht]
  \vspace{.25em}
\begin{lstlisting}
% OLON rule (PRIMARY obligation to return car)
fail_to_return_car :-  friend(J,X), car(C),
    borrowed_car(J,X,C,Tb), 
    not ok_car_returned(J,X,C),
    not fail_to_return_car.
ok_car_returned(J,X,C) :-
    borrowed_car(J,X,C,Tb),
    car_returned(J,X,C,Tr), Tr .>. Tb.
% OLON rule (secondary obligation: return by noon)
fail_to_return_by_noon :- friend(J,X), car(C),
    car_returned(J,X,C,Tr), Tr .>. 12,
    borrowed_car(J,X,C,Tb), Tr .>. Tb,
    not fail_to_return_by_noon.
% OLON rule for secondary obligation
%          to return with battery level ok
fail_to_return_ok_battery :-
    friend(J,X), car(C),
    borrowed_car(J, X, C, Tb), Tr .>. Tb,  %
    car_returned(J, X, C, Tr), 
    not battery_ok_to_return(C, Tb, Tr), 
    -ignore_battery_level(C),         %model permissibility
    not fail_to_return_ok_battery.
battery_ok_to_return(C, Tb, Tr). :- 
    same_battery_level(C, Tb, Tr).
same_battery_level(C, T1, T2) :- car(C), T2 .>. T1,
    batterylvl(C, T1, L1), batterylvl(C, T2, L2),
    D .=. L1 - L2, D .<. 0.05*L1.
%Model permissibility
-ignore_battery_level(C) :- not needBatteryChange(C).
% if the car is wrecked, John can't return the car
fail_to_return_car :- friend(J,X), car(C),
    borrowed_car(J,X,C,Tb), wrecked(C,Tw), Tw .>. Tb.
% fail to return before noon if become sick
fail_to_return_by_noon :- friend(J,X), car(C),
    borrowed_car(J,X,C,Tb), sick(J, Ts), 
    Ts .>. Tb, Ts .=<. 12.
% ... also if fail to return car
fail_to_return_by_noon :- fail_to_return_car. 
% if no money, can't charge depleted battery
fail_to_return_ok_battery :- friend(J,X), car(C),
    borrowed_car(J,X,C,Tb), financially_broke(J).
% ... also if fail to return car
fail_to_return_ok_battery :- fail_to_return_car.
\end{lstlisting}
  \caption{s(CASP) Encoding of Car-borrowing Example}
 \label{fig:car}
 \vspace{-1em}
\end{figure}

We further explain the above code using various narratives. We explain the version with continuous time (Fig.~\ref{fig:car}). The reviewer is encouraged to try out them: \hfill \scasp{car}

\begin{description}

\item [Narrative 1:] wrecked the car: couldn't keep any obligation.

\begin{lstlisting}
wrecked(smith_bmw, 4).
borrowed_car(jones, smith, smith_bmw, 0).
\end{lstlisting}

\noindent
Since the obligations are encoded as global constraints, s(CASP) checks them independently of the query. Thus, by simply invoking the query \code{?-true}, the evaluation checks if each narrative meets all the applicable obligations and if any are violated.
such as in this narrative in which the car cannot be returned because it is wrecked (lines 38-40 in Fig.~\ref{fig:car}).

\smallskip
\item [Narrative 2:] car returned before noon with the same battery level. 

\begin{lstlisting}
car_returned(jones, smith, smith_bmw, 5). 
borrowed_car(jones, smith, smith_bmw, 0).
batterylvl(smith_bmw, 0, 200).
batterylvl(smith_bmw, 5, 200).
\end{lstlisting}

\noindent
Given this narrative, all the obligations are met.

\smallskip
\item [Narrative 3:] car returned late after 12 due to being sick, but battery level is ok.

\begin{lstlisting}
car_returned(jones, smith, smith_bmw, 14).
sick(jones,8).     
borrowed_car(jones, smith, smith_bmw, 0).
batterylvl(smith_bmw, 0, 200).
batterylvl(smith_bmw, 14, 195).
\end{lstlisting}

\noindent
The evaluation of this narrative succeeds because \code{fail_to_return_by_noon} is supported by the fact of being sick (see lines 42-25 in Fig.~\ref{fig:car}). This information is shown by s(CASP) in the partial model and in the justifications tree (for the reader's convenience, they are available in \ref{sec:a1}).

\smallskip
\item [Narrative 4:] car returned before noon, but with a low battery (no money to get it charged).

\begin{lstlisting}
borrowed_car(jones, smith, smith_bmw, 0).
batterylvl(smith_bmw, 0, 200).
batterylvl(smith_bmw, 10, 150).
car_returned(jones, smith, smith_bmw, 10).
financially_broke(jones).
\end{lstlisting}

  \noindent
  Under this narrative, the exception \code{fail_to_return_ok_battery} holds because the fact \code{financially_broke(jones)} appears in the narrative, i.e., Jones is financially broke.

\smallskip
\item [Narrative 5:] car returned but late and with low battery levels due to being too broke to get it charged.

\begin{lstlisting}
borrowed_car(jones, smith,smith_bmw,0).
batterylvl(smith_bmw, 0, 200).
batterylvl(smith_bmw, 14, 150).
car_returned(jones, smith, smith_bmw, 14).
sick(jones, 10).
financially_broke(jones).
\end{lstlisting}

  \noindent
  Under this narrative, the exceptions \code{fail_to_return_by_noon} and \code{fail_to_return_ok_battery} holds. The first one because he is sick, and the second one because he is financially broke.

\smallskip
\item [Narrative 6:] car returned before noon, low battery levels, but battery needs replacement (being broke should have no impact)

\begin{lstlisting}
borrowed_car(jones, smith, smith_bmw, 0).
batterylvl(smith_bmw, 0, 200).
batterylvl(smith_bmw, 10, 150).
car_returned(jones, smith, smith_bmw, 10).
needBatteryChange(smith_bmw).
financially_broke(jones).  
\end{lstlisting}

  \noindent
  In this scenario, it is possible to prove \code{fail_to_return_ok_battery} because he is financially broke, but since the battery needs replacement, this exception is not necessary (see the following example).
     
\smallskip
\item[Narrative 7:] car returned before noon, low battery levels, but battery needs replacement (however, Jones is not financially broke)

\begin{lstlisting}
borrowed_car(jones, smith, smith_bmw, 0).
batterylvl(smith_bmw, 0, 200).
batterylvl(smith_bmw, 10, 150).
car_returned(jones, smith, smith_bmw, 10).
needBatteryChange(smith_bmw).
-financially_broke(jones).  
\end{lstlisting}

  \noindent
  Note that we are using classical negation, i.e., \code{-financially_broke(jones)} to indicate that we have evidence that he is not broke, and now the evaluation succeeds without exceptions because in this narrative we also have that the battery needs replacement, and this exception is defined in line 11 of Fig.~\ref{fig:car} and in lines 8-9 as an appropriate exception (using the default negation), i.e., \code{not abnormal_battery_status(C)}.

\smallskip
\item[Narrative 8:] car returned before noon,  battery level is the same, but battery needs replacement. We show that in the situation where battery needs replacement (it's too old), but battery charge is the same, the need to change battery does not affect returning of the car on time. 

\begin{lstlisting}
borrowed_car(jones, smith, smith_bmw, 0).
batterylvl(smith_bmw, 0, 200).
batterylvl(smith_bmw, 10, 150).
car_returned(jones, smith, smith_bmw, 10).
needBatteryChange(smith_bmw). 
\end{lstlisting}

  \noindent
  Note that we are using classical negation, i.e., \code{-financially_broke(jones)} to indicate that we have evidence that he is not broke, and now the evaluation succeeds without exceptions because in this narrative we also have that the battery needs replacement, and this exception is defined in line 11 of Fig.~\ref{fig:car} and in lines 8-9 as an appropriate exception (using the default negation), i.e., \code{not abnormal_battery_status(C)}.

\end{description}

 \newpage

\redsection
\section{Partial Models and Justifications}
\label{sec:a1}

As we already mentioned s(CASP) evaluation provides partial models (including only the literals needed to support the query) and a justification tree that can be translated into a human-readable explanation (see Reference [1] below) for details.

To make the model more readable let's include the directive \code{#show fail_to_return_car/0, fail_to_return_by_noon/0, fail_to_return_ok_battery/0, fail_to_return_car/0, fail_to_return_ok_battery/0} so only the exceptions are displayed.
For example, given the Narrative 3, presented in Section~\ref{sec:carexample}, the resulting partial model is:

\begin{lstlisting}
{ fail_to_return_by_noon }
\end{lstlisting}

And it corresponding justification tree is:

\begin{lstlisting}[style=tree]
JUSTIFICATION_TREE:
true,
global_constraint :-
    not o_chk_1 :-
        not friend(Var0 | {Var0 \= jones},Var1),
        not friend(jones,Var2 | {Var2 \= smith}),
        friend(jones,smith),
        not car(Var3 | {Var3 \= smith_bmw}),
        proved(friend(jones,smith)),
        car(smith_bmw),
        not car_returned(jones,smith,
                        smith_bmw,Var4 | {Var4 \= 14}),
        proved(car(smith_bmw)),
        car_returned(jones,smith,smith_bmw,14),
        not borrowed_car(jones,smith,
                         smith_bmw,Var5 | {Var5 \= 0}),
        proved(car_returned(jones,smith,smith_bmw,14)),
        borrowed_car(jones,smith,smith_bmw,0),
        fail_to_return_by_noon :-
            friend(jones,smith),
            car(smith_bmw),
            borrowed_car(jones,smith,smith_bmw,0),
            sick(jones,8).
    not o_chk_2 :-
        not friend(Var6 | {Var6 \= jones},Var7),
        not friend(jones,Var8 | {Var8 \= smith}),
        not car(Var9 | {Var9 \= smith_bmw}),
        not borrowed_car(jones,smith,
                       smith_bmw,Var10 | {Var10 \= 0}),
        proved(borrowed_car(jones,smith,smith_bmw,0)),
        not sick(jones,Var11 | {Var11 \= 8}),
        proved(sick(jones,8)),
        proved(fail_to_return_by_noon).
    not o_chk_3 :-
        not friend(Var12 | {Var12 \= jones},Var13),
        not friend(jones,Var14 | {Var14 \= smith}),
        proved(friend(jones,smith)),
        not car(Var15 | {Var15 \= smith_bmw}),
        proved(car(smith_bmw)),
        not borrowed_car(jones,smith,
                       smith_bmw,Var16 | {Var16 \= 0}),
        proved(borrowed_car(jones,smith,smith_bmw,0)),
        ok_car_returned(jones,smith,smith_bmw) :-
            car_returned(jones,smith,smith_bmw,14).
    not o_chk_4 :-
        not friend(Var17 | {Var17 \= jones},Var18),
        not friend(jones,Var19 | {Var19 \= smith}),
        not car(Var20 | {Var20 \= smith_bmw}),
        not borrowed_car(jones,smith,
                       smith_bmw,Var21 | {Var21 \= 0}),
        not car_returned(jones,smith,
                      smith_bmw,Var22 | {Var22 #> 14}),
        not car_returned(jones,smith,
           smith_bmw,Var23 | {Var23 #> 0,Var23 #< 14}),
        proved(car_returned(jones,smith,smith_bmw,14)),
        battery_ok_to_return(smith_bmw,0,14) :-
            same_battery_level(smith_bmw,0,14) :-
                batterylvl(smith_bmw,0,200),
                batterylvl(smith_bmw,14,195),
                diff(200,195,5).
            not abnormal_battery_status(smith_bmw) :-
                proved(car(smith_bmw)),
                not needBatteryChange(smith_bmw).
    not o_chk_5 :-
        not friend(Var24 | {Var24 \= jones},Var25),
        not friend(jones,Var26 | {Var26 \= smith}),
        not car(Var27 | {Var27 \= smith_bmw}),
        not borrowed_car(jones,smith,
                       smith_bmw,Var28 | {Var28 \= 0}),
        not financially_broke(jones).
\end{lstlisting}

\noindent
In this justification tree, we can observe that all five OLON rules are satisfied, but let's focus in particular on the first check rule (lines 4-21), where the support for \code{fail_to_return_by_noon} is justified (lines 17-21), since, as we can see in line 21 the fact \code{sick(jones,8)} holds, i.e., this narrative asserts that Jones fell ill at 8 and therefore, he cannot return the car before noon. Note that this justification corresponds to the clause for \code{fail_to_return_by_noon} in lines 42-45 of Fig.~\ref{fig:car}.

\medskip 
\noindent \hangindent 0.5in [1] Arias, J., Carro, M., Chen, Z., and Gupta, G. Justifications for goal-directed constraint answer set programming. In ICLP-TC 2020, volume 325, 59–72. EPTCS.

\end{document}